\documentclass[letterpaper]{article} 
\usepackage{aaai24}  
\usepackage{times}  
\usepackage{helvet}  
\usepackage{courier}  
\usepackage[hyphens]{url}  
\usepackage{graphicx} 
\usepackage{natbib}  
\usepackage{caption} 
\usepackage{algorithm}
\usepackage{algorithmic}
\usepackage{multirow}
\usepackage{booktabs}
\usepackage{amssymb}
\usepackage{amsmath}
\usepackage{newfloat}
\usepackage{listings}
\DeclareCaptionStyle{ruled}{labelfont=normalfont,labelsep=colon,strut=off} 
\floatstyle{ruled}
\newfloat{listing}{tb}{lst}{}
\floatname{listing}{Listing}
\title{Learning Continuous Implicit Field with Local Distance Indicator for Arbitrary-Scale Point Cloud Upsampling}
\author{
    Shujuan Li\textsuperscript{\rm 1}\equalcontrib, 
    Junsheng Zhou\textsuperscript{\rm 1}\equalcontrib, 
    Baorui Ma\textsuperscript{\rm 1,2},
    Yu-Shen Liu\textsuperscript{\rm 1}\thanks{Corresponding author: Yu-Shen Liu}, 
    Zhizhong Han\textsuperscript{\rm 3}
}
\affiliations{
    \textsuperscript{\rm 1}School of Software, Tsinghua University, Beijing, China,
    \textsuperscript{\rm 2}Beijing Academy of Artificial Intelligence\\
    \textsuperscript{\rm 3}Department of Computer Science, Wayne State University, Detroit, USA\\
    {lisj22, zhoujs21}@mails.tsinghua.edu.cn,
    brma@baai.ac.cn,
    liuyushen@tsinghua.edu.cn, h312h@wayne.edu
}

\usepackage{bibentry}

\begin{document}

\maketitle
\begin{abstract}
Point cloud upsampling aims to generate dense and uniformly distributed point sets from a sparse point cloud, which plays a critical role in 3D computer vision. Previous methods typically split a sparse point cloud into several local patches, upsample patch points, and merge all upsampled patches. However, these methods often produce holes, outliers or non-uniformity due to the splitting and merging process which does not maintain consistency among local patches.
To address these issues, we propose a novel approach that learns an unsigned distance field guided by local priors for point cloud upsampling. Specifically, we train a local distance indicator (LDI) that predicts the unsigned distance from a query point to a local implicit surface. Utilizing the learned LDI, we learn an unsigned distance field to represent the sparse point cloud with patch consistency. At inference time, we randomly sample queries around the sparse point cloud, and project these query points onto the zero-level set of the learned implicit field to generate a dense point cloud. 
We justify that the implicit field is naturally continuous, which inherently enables the application of arbitrary-scale upsampling without necessarily retraining for various scales. We conduct comprehensive experiments on both synthetic data and real scans, and report state-of-the-art results under widely used benchmarks. Project page: \url{https://lisj575.github.io/APU-LDI}
\end{abstract}

\begin{figure}
\centering
\includegraphics[width=0.95\linewidth]{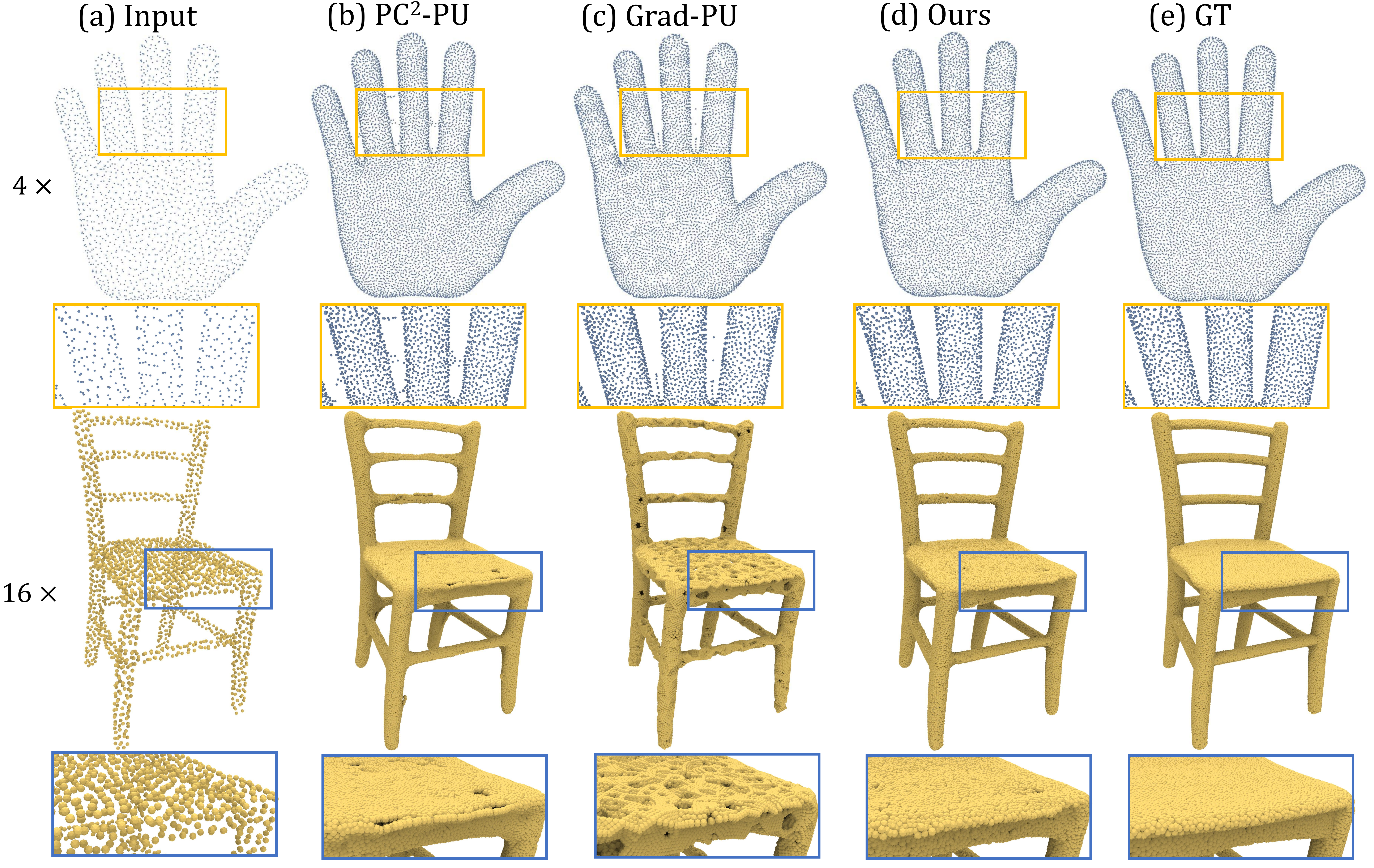}
\caption{Comparison with PC$^2$-PU \cite{long2022pc2} and Grad-PU \cite{he2023grad}. The input point cloud has 2048 points, and ground truth has 8192 points for the palm and 32768 points for the chair. Previous methods upsample the point cloud at the local level, which results in outliers, holes and non-uniform when combining local patches. We represent the entire point cloud as a continuous implicit field and generate globally consistent results.}
\label{fig:methods_cpmparison}
\end{figure}

\section{Introduction}
Point cloud, as a lightweight representation of 3D objects and scenes, has gained significant attention. It can be easily obtained through depth cameras or other 3D scanning devices. However, the raw point clouds obtained from these devices are often sparse and non-uniform, which negatively impacts various downstream applications like semantic understanding \cite{zhang2019review}, surface reconstruction \cite{chen2023unsupervised}, and point cloud rendering \cite{dai2020neural}. Hence, it is crucial to generate dense and uniformly distributed points from the raw point cloud data.

Recently, the learning-based methods on point cloud upsampling have achieved promising results. The mainstream methods typically split a sparse point cloud into several patches, in which a neural network is trained for predicting surface signals, such as surface points \cite{yu2018pu, li2019pu, qian2021pu, feng2022neural}, offsets \cite{li2021point}, interpolation weights \cite{luo2021pu, qian2021deep}, and the point-to-surface distances \cite{he2023grad, zhao2023self}. With the help of learned signals, points on each patch become denser. Finally, the upsampled patches are combined into a complete point cloud as the final prediction. However, the inconsistency among local patches causes two issues as shown in Figure \ref{fig:methods_cpmparison}. First, the geometry of the patch boundary is hard to determine without the global shape information, resulting in outliers on the final dense point cloud. Second, combining inconsistent patches leads to holes or non-uniformity on the final point cloud.

To solve these issues, we propose to learn an unsigned distance field to represent a shape with the guidance from a pre-trained local distance indicator (LDI) learned from dense point cloud patches at local level. With the continuous implicit field and local distance prior, we produce accurate upsampling results which capture rich geometric details and also retain uniformity due to the global consistency of the entire shape. Specifically, we first train a LDI that predicts the distance from a query location to the underlying surface, which aims to learn a sparse-to-dense prior on local patches.

To predict the distance more accurately and robustly, we further design an attention-based module to learn attention weights between the query point and its neighbouring patch points at the feature space. Given the learned LDI, we then utilize it as the supervision to guide the learning of the global implicit function which represents the underlying surface of the entire sparse point cloud. Finally, for randomly sampled query points, we use the implicit function to obtain the distances and the gradients, and then project queries onto a surface by pulling them along the direction of gradients at a stride of distances to form the dense point cloud. By introducing the continuous implicit function, we achieve an accurate and uniform point cloud upsampling with consistency among patches. We further justify that the continuity of the implicit surface can control the density of the dense point cloud by changing the number of sampled query points, which enables our method to achieve arbitrary-scale upsampling without retraining for various scales.

Our contributions are summarized as follows.
\begin{itemize}
\item We propose to learn an unsigned distance field from a sparse point cloud with the guidance of a pre-trained distance indicator at local level. By pulling 3D queries into the zero level set of the field, we can achieve point cloud upsampling at arbitrary-scales.

\item We design an attention-based module to learn a local distance indicator that estimates point-to-surface distances, which brings rich local priors to the global implicit field.

\item We achieve state-of-the-art results in arbitrary-scale point cloud upsampling for synthetic data and real scans on widely used benchmarks.

\end{itemize}

\section{Related Work}
\subsection{Optimization-based Point Cloud Upsampling}
Optimization-based methods generate dense point clouds by employing well-designed shape priors. As the pioneering work in point cloud upsampling, Alexa et al. \citeyear{alexa2003computing} propose to add points at the vertices of Voronoi diagrams. To preserve more geometrical structures, Huang et al. \citeyear{huang2009consolidation} introduce weighted locally optimal projection operator and robust normal estimation process for point cloud upsampling. Later, Edge-Aware Resampling (EAR) \cite{huang2013edge} achieves fine detail preservation by resampling points away from edges and progressively upsampling them to approach the edge singularities. In addition, several methods are proposed to deal with different conditions such as noisy data \cite{lipman2007parameterization} and missing regions \cite{wu2015deep}. 
Overall, optimization-based methods heavily rely on surface smoothness assumptions, which often leads to struggles between geometric detail and robustness.
\subsection{Learning-based Point Cloud Upsampling}

Recently, with the development of deep learning, learning-based methods \cite{yu2018pu, qian2020pugeo, luo2021pu, liu2022spu, feng2022neural} are proposed for point cloud upsampling. Yu et al. \cite{yu2018pu} propose the first learning-based point cloud upsampling method PU-Net. PU-GAN \cite{li2019pu} introduces generative adversarial network \cite{goodfellow2014generative} to generate uniform point clouds. PU-GCN \cite{qian2021pu} uses a Graph Convolutional Network to learn better local features. Considering the patch connection, PC${^2}$-PU \cite{long2022pc2} introduces patch-to-patch and point-to-point modules to help the combination of different patches. Meta-PU \cite{ye2021meta} supports arbitrary scale factor upsampling using meta-network to adjust the weights of residual graph convolution blocks dynamically. PU-SMG \cite{dell2022arbitrary} maps the sparse point cloud to the Spherical Mixture of Gaussians. More recently, Grad-PU \cite{he2023grad} samples more accurate point clouds by predicting point-to-point distances. PU-SSAS \cite{zhao2023self} uses implicit neural functions to estimate projection distance and direction with unsigned distance and normal supervision. Existing methods focus on the geometrical structure at the local level and use networks to learn rich patch information, but the ambiguous boundaries at local patches always lead to issues like holes, outliers and non-uniformity. Although PC${^2}$-PU tries to alleviate the connection problem between patches with patch correlation module, they still suffer from outliers for the lack of global consistency of the global shape. We propose to learn an unsigned distance field of the entire shape with local priors, which introduces globally consistent information.

\begin{figure}
\centering
\includegraphics[width=0.95\linewidth]{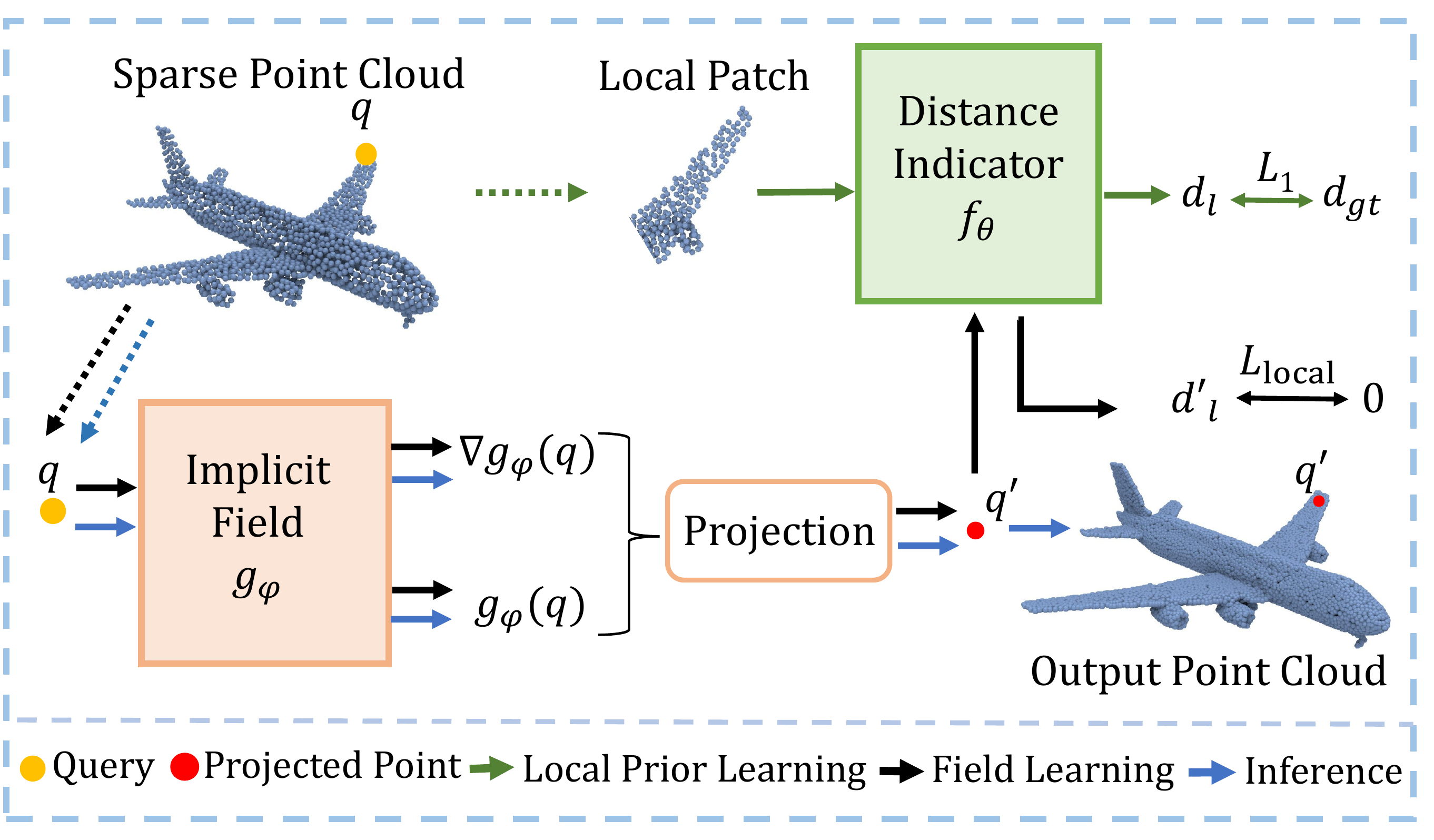}
\caption{Overview of the proposed method. The key idea of our method is to learn an unsigned distance field to represent the sparse point cloud at global level with the guidance from a pre-trained LDI learned from dense local patches.}
\label{fig:overview}
\end{figure} 

\subsection{Implicit Neural Representation}
Implicit neural representation has shown great advantages in continuous surface representation \cite{zhang2023fast, ma2023towards}. Usually, the implicit function is learned by a neural network, the input of the network is a query point, and the output is occupancy probability \cite{mescheder2019occupancy, chen2019learning} or signed/unsigned distance \cite{park2019deepsdf, michalkiewicz2019implicit}. Moreover, Neural Radiance Fields (NeRFs) \cite{mildenhall2021nerf} use implicit representations to encode the geometry and appearance information. Recently, some researches \cite{ ma2022reconstructing, zhou2023levelset, jin2023multi, chen2023gridpull} propose to learn implicit continuous surfaces from 3D point clouds. Among them, OnSurfacePrior \cite{ma2022reconstructing} proposes to learn surface priors to guide the learning of signed distance functions (SDF). However, directly transferring this method to the point cloud upsampling task leads to oversmoothed surfaces due to the MLP-based prior network and the lack of sufficient global constraints. In our method, the designed attention-based distance indicator guides the neural network to learn a global unsigned distance field that handles more complex shapes, and the additional global constraints help to preserve more details.
More recently, some methods \cite{zhao2022self, zhao2023self} have achieved encouraging results by applying implicit representations to point cloud upsampling tasks, while these methods often require other shape information, such as normals and unsigned distances. Our method only requires dense point clouds as supervision.

\section{Methodology}
\subsection{Overview}
Given the input sparse point cloud $S=\{s_i\}^N_{i=1}$, point cloud upsampling aims to generate a dense point cloud $T=\{t_j\}_{j=1}^M$, where $M=rN$ and $r$ is the scale factor. The generated point cloud should meet the requirement that each point of the dense point cloud is located at the underlying surface represented by the sparse point cloud and the dense point cloud is uniformly distributed.
The overview of our proposed method is shown in Figure \ref{fig:overview}. We learn an unsigned distance field to represent the continuous surface described by the sparse point cloud $S$, which is guided by the local distance indicator that predicts the distance from the query point to the local patch. At inference time, we project the query points to the zero-level set of the implicit field to generate the dense point cloud. In this Section, we will begin by introducing the specifics of the local distance indicator. Subsequently, we will delve into the learning process of the global implicit field, followed by the technique of arbitrary-scale upsampling on the continuous implicit field.
\subsection{Local Distance Indicator (LDI)}

To learn more accurate geometrical details, we first train a neural network as a local distance indicator to learn upsampling priors from grouped patches in local level. For a sparse point cloud $S$, we first split it into different patches like previous methods \cite{long2022pc2, he2023grad}. Then, we sample queries around each patch point with normal distribution. The distance indicator takes the query point $q$ and its corresponding local patch $P$ as the input and predicts the distance $d_l$ from the query point to the local patch in a local coordinate system as follows:
\begin{figure}
\centering
\includegraphics[width=1.0\linewidth]{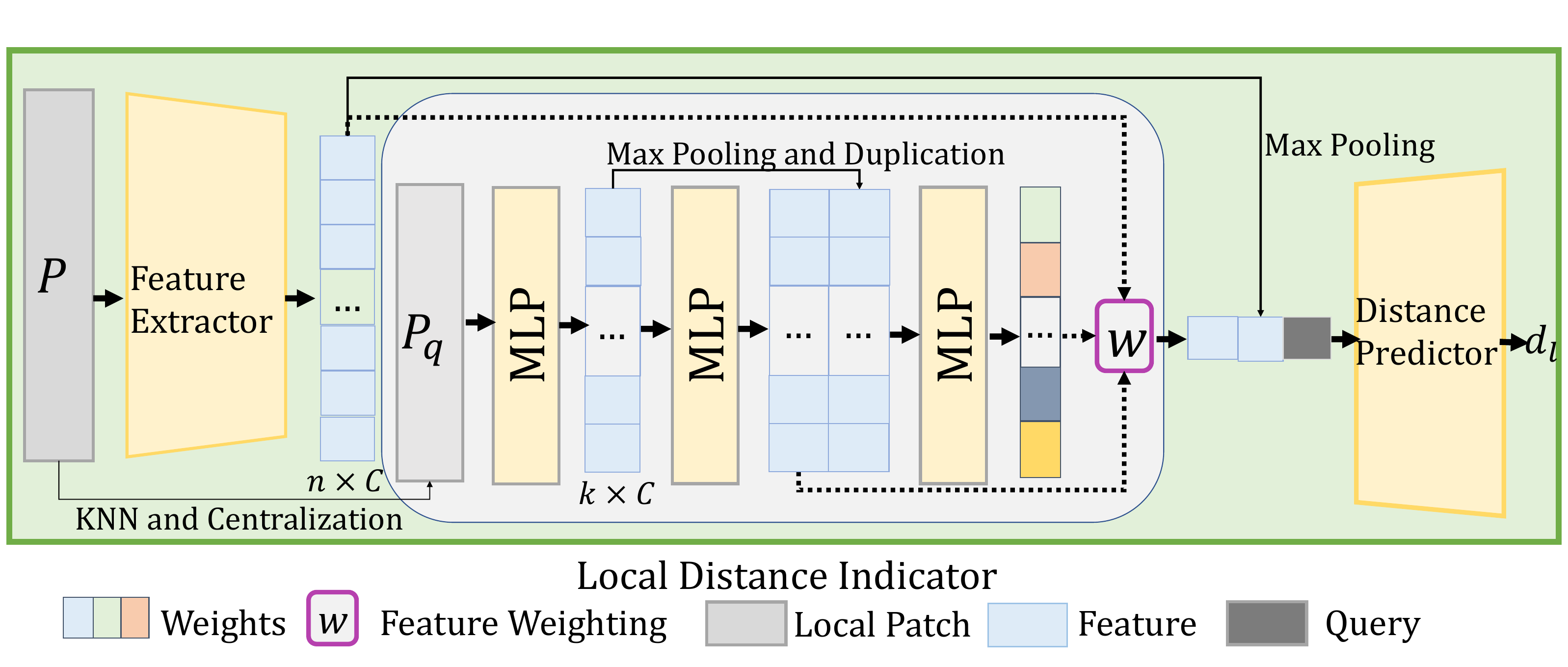}
\caption{Structure of the local distance indicator. The attention-based distance indicator predicts the distance between the query point and the patch surface.}
\label{fig:local_indicator}
\end{figure}

\begin{equation}
\label{eq:definition of distance indicator}
f_{\theta}(q,P)=d_l,
\end{equation}
where $f_{\theta}$ denotes the network parameterized by $\theta$.

To predict the distance more accurately and robustly, we further design an attention-based module to learn attention weights between the query point and its neighboring points at the feature space. The structure is shown in Figure \ref{fig:local_indicator}. Specifically, for patch $P=\{p_h\}_{h=1}^n$, we use the feature extractor to obtain the point-wise patch features $F=\{f_h\}_{h=1}^n$ for patch points. Then we find a smaller region $P_q = \{p_l\}_{l=1}^k$ where $k < n$ with k-nearest neighbor algorithm (KNN) for query point $q$. We further normalize $P_q$ by centering it to the query point $q$ followed by MLPs to extract the relative features $F'=\{f'_d\}_{d=1}^k$. To add the global information of the patch to the features, we use max-pooling on the relative features to generate the global feature. Finally, we concatenate the relative features and global features together and predict the relative weights $w_d$ between the query point and the neighboring points. With the relative weights, we find the most important features on the local patch. We also use the relative features of $q$ and the patch points to add relative distance perception for more accurate prediction. The final feature $f_q$ for $q$ is computed by weighting patch features and relative features:
\begin{equation}
\label{eq:final feature for query point}
f_q = \sum_{d=1}^k{w_d\cdot f'_d+(1-w_d) \cdot f_d}.
\end{equation}
Then, we concatenate $f_q$, the global feature of patch features and the query point coordinate together and use a distance predictor to map the final vector to point-to-surface distance $d_l$. We train the network by minimizing the L1 loss between the predicted distance and the distance $d_{gt}$ between the query point and its nearest point in the dense point cloud:
\begin{equation}
\label{eq:local_distance_indicator}
L_1 = |d_l-d_{gt}|.
\end{equation}

\begin{figure}
    \centering
    \includegraphics[width=1.0\linewidth]{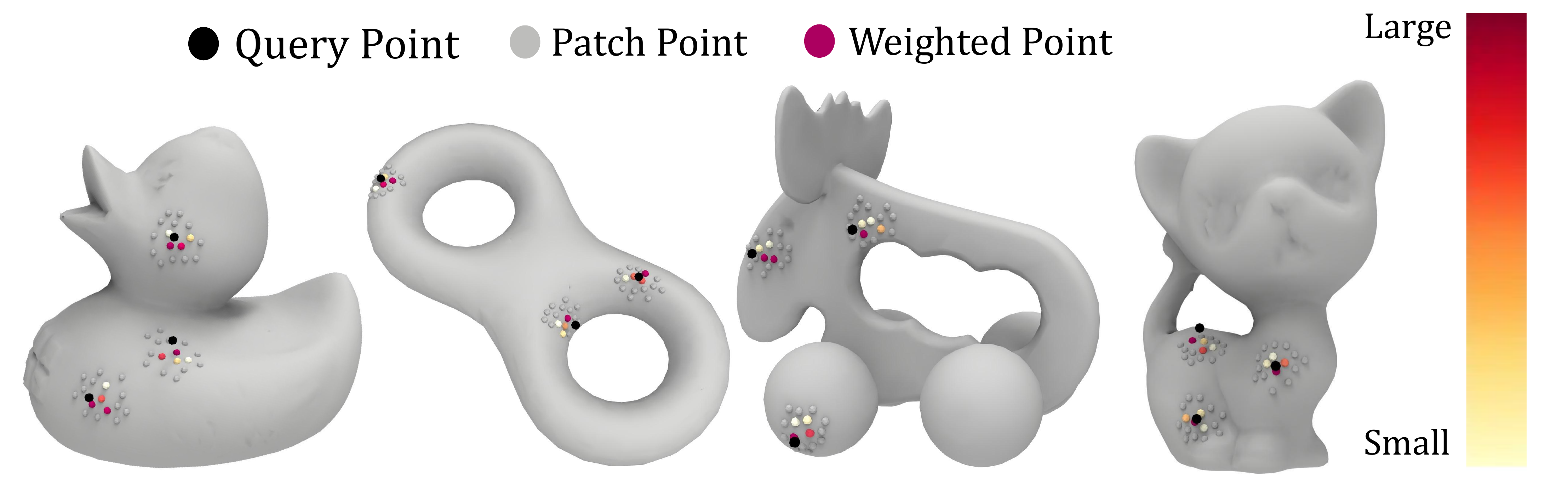}
    \caption{The visualization of learned weights.}
    \label{fig:weight}
\end{figure}

We visualize the weights learned by the attention module in Figure \ref{fig:weight}. For a query point, we select its four neighbors to generate its final feature. The color map indicates the values of the weights, where the darker the color, the greater the weight. The results show that the closer points have relatively larger weights, which proves that the learned weights are reasonable. In addition, the value of the weights are not strictly positively correlated with the distances, which proves that the attention module is adaptable.

\subsection{Learning the Continuous Field}

With the learned LDI, we can learn the global implicit field to represent the entire point cloud. As shown in Figure \ref{fig:overview}, for the query point $q$, we use the neural network $g_\phi$ that is parameterized by $\phi$ to predict its distance $d_g$ to the surface in the global coordinate system. Here, we use the pulling process introduced by \cite{ma2021neural} to project the query point $q$ to the surface point $q'$, formulated as,
\begin{equation}
\label{eq:pulling}
q'=q-{g_\phi(q)}\nabla{g_\phi(q)}/||{g_\phi}(q)||_2,
\end{equation}
where $\nabla{g_\phi(q)}$ is the gradient at $q$ within the network. We aim to learn to project $q_i$ to the underlying surface represented by the sparse point cloud $S$. However, the real position of $q'$ is unknown due to the sparsity of $S$, where we leverage our learned LDI as supervision for guiding the implicit field learning process.

Specifically, we use KNN to find the local patch $P_{q'}$ that $q'$ is centered at and use the fixed indicator to predict the distance $f_\theta(q', P_{q'})$ from $q'$ to $P_{q'}$ at the local level. We justify that the indicated distance $f_\theta(q', P_{q'})$ of $q'$ increases as $q'$ gets projected to locations far from the underlying surface and decreases when $q'$ is projected close to the underlying surface. Additionally, $f_\theta(q', P_{q'}) = 0$ means that $q'$ is projected to the correct position. Therefore, we minimize the L1 loss between the predicted distance and 0:
\begin{equation}
\label{eq:local_loss}
L_{\rm{local}} = |f_\theta(q', P_{q'})|.
\end{equation}

To take full advantage of the information of the global point cloud, we also add some additional constraints. 

At the early stage of training, the global network $g_{\phi}$ projects the query point to the location away from the surface, leading to large $f_{\theta}(q', P_{q'})$, which makes the optimization difficult. Therefore, we introduce the surface point constraint that encourages $q'$ to get projected closer to its nearest point $q_s$ on the sparse point cloud:
\begin{equation}
\label{eq:np_loss}
L_{\rm{np}} = ||q'-q_s||.
\end{equation}

To make the implicit field faithfully describe the underlying surface described by the sparse point cloud, we add a term to encourage the zero-level set of the implicit field to pass through the input points $\{s_i\}_{i=1}^N$. The loss term is:
\begin{equation}
\label{eq:np_surf}
L_{\rm{surf}} = |g_\phi(s_i)|.
\end{equation}

Due to the approximation error of the point-to-surface distance, the gradient of the implicit field is not the exact normal direction, leading to errors when $q'$ is very close to the underlying surface. To learn a more accurate distance and direction of the query point $q$, we add a shortest path constraint introduced by \cite{ma2022reconstructing}:
\begin{equation}
\label{eq:sp_loss}
L_{\rm{sp}} = |g_\phi(q)|.
\end{equation}

The total global loss is given as:
\begin{equation}
\label{eq:global_loss}
L_{\rm{global}} = \alpha \cdot L_{\rm{np}} + \beta \cdot L_{\rm{surf}} + \gamma \cdot L_{\rm{sp}},
\end{equation}
where $\alpha$, $\beta$ and $\gamma$ are the weights for different loss terms.

Therefore, the total loss is $L=L_{local}+L_{global}$. After training, the global implicit field can be leveraged to predict the projection direction and distance for an arbitrary query point to reach the underlying surface. The advantage of the implicit function is that the neural network learns the relation between the spatial point and the underlying surface, so it can represent the shape with arbitrary precision theoretically, which leads to a continuous surface representation. Next, we show the details of arbitrary-scale point cloud upsampling with the implicit field.

\subsection{Arbitrary-Scale Point Cloud Upsampling}
With the learned global implicit field, each query point can be projected to the continuous underlying surface. We change the number of query points to achieve arbitray-scale upsampling without retraining. For the surface point $s \in S$, we sample offsets from a uniform distribution that is bounded by zero and the nearest distance of $s$ and other points in $S$. Then, we add the offsets to $s$ to form the queries. We use the implicit filed to predict the gradient and distance for each query point, and project it to the surface by pulling $q$ along gradient $\nabla{g_\phi(q)}$ at a stride of the predicted distance $g_\phi(q)$ with Eq. \ref{eq:pulling}. To make the upsampled point cloud more uniform, we generate more query points than the target and employ farthest points sampling (FPS) \cite{eldar1997farthest} algorithm to downsample it to the target number.

\subsection{Implementation Details}
For the local distance indicator, the input patches are normalized and interpolated using the same strategies in \cite{he2023grad}. During the learning of global implicit field, we adopt the similar strategy as \cite{zhou2022learning} to sample about 240 queries around each point $s_i$. Specifically, we add Gaussian noises to the sparse point cloud to generate query points where the mean is $0$ and the standard variance is $20\%$ of the distance between $s_i$ and its 50-th nearest point on the point cloud. We also add some query points randomly sampled on the whole space.

We employ network structures of encoder and decoder similar to \cite{he2023grad} for the feature extractor and distance predictor individually. The encoder mainly consists of several dense blocks \cite{huang2017densely} and the decoder contains several MLP layers with ReLU activation. Besides, we adopt a neural network
similar to DeepSDF \cite{park2019deepsdf} to represent the global implicit field, which is composed of 8 fully connected layers with a residual connection.

\section{Experimental Results}

\begin{figure*}
\centering
\includegraphics[width=1.0\linewidth]{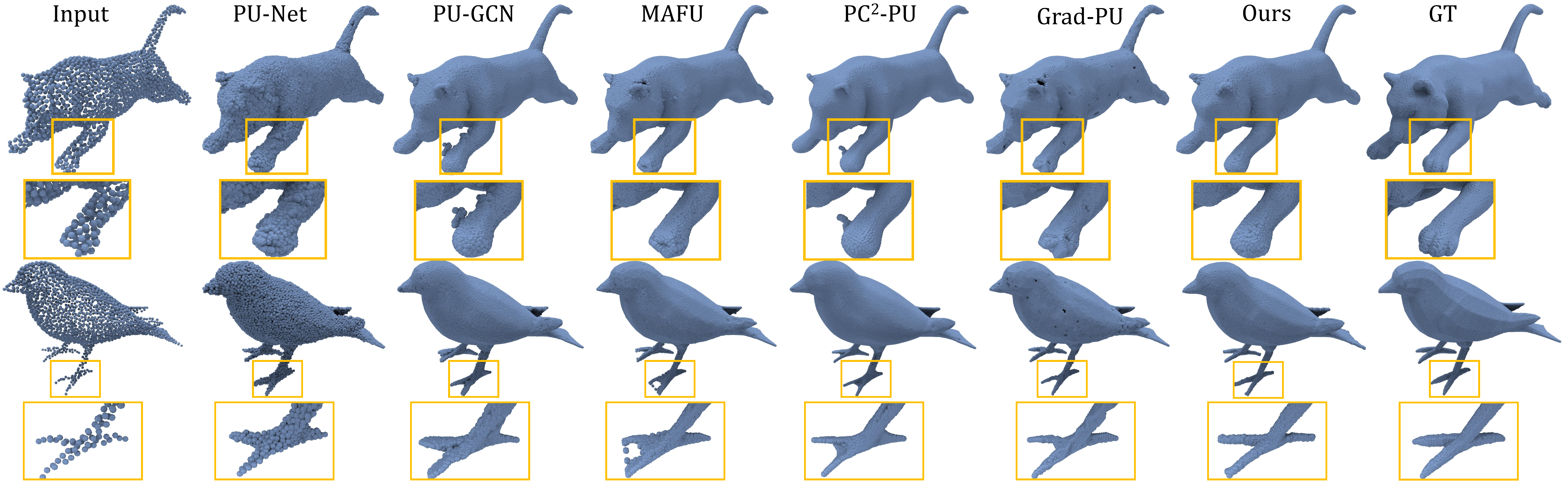}
\caption{Qualitative comparisons of $16\times$ upsampling results on the the PU-GAN dataset. Our method generates uniform dense point clouds with fewer holes and outliers than state-of-the-art methods.}
\label{fig:pugan_compare}
\end{figure*}
\subsection{Experiment Setup}
\subsubsection{Dataset.}
We adopt two public synthetic datasets to evaluate our method. First, we make a comparison with the baseline methods on the PU-GAN dataset provided by \cite{li2019pu}. This dataset contains 147 objects with various geometries, and 120 objects are used for training and the rest are used for testing. Second, we adopt a more complex dataset PU1K provided by \cite{qian2021pu}, which contains 1,147 3D models collected from different categories, and 1020 objects are used for training. We use the patch to form the training data provided by the authors, which contains the pairs of sparse patches with 256 points and dense patches with 1024 points. We use the sample data augmentation strategy like previous methods \cite{he2023grad} to make data diverse, including normalizing, rotation and random scale. For testing, each input point cloud generated from the watertight meshes contains 2048 points and ground truth contains 8096 points. We use the training sets of the two datasets to learn the local priors, and then learn the implicit field on the sparse point cloud of the test. All experiments are conducted on a single GeForce RTX 3090 gpu. We set $\alpha=1.0, \gamma=0.1$, and $\beta$ decays linearly from 0.5 to 0.

\begin{table}
    \small    
    \centering
    \begin{tabular}{lcccccc}  
    \toprule 
    {Factor}&\multicolumn{3}{c}{$4\times$ (r=4)}   &\multicolumn{3}{c}{$16\times$ (r=16)}\cr
    \cmidrule{2-4} \cmidrule{5-7}
    \multirow{2}{*}{Methods} &CD$\downarrow$& HD$\downarrow$& P2F$\downarrow$ & CD$\downarrow$& HD$\downarrow$& P2F$\downarrow$ \cr
    & $10^{-3}$ & $10^{-3}$ & $10^{-3}$ & $10^{-3}$ & $10^{-3}$ & $10^{-3}$ \cr
    \midrule
    PU-Net&0.507&4.312&4.694&0.596&6.929&6.014 \cr
    MPU &0.289&4.771&2.830&0.175&5.898&3.111 \cr
    PU-GAN &0.284&3.301&2.660&0.202&5.045&2.996 \cr
    Dis-PU &0.278&3.447&2.343&0.180&4.888&2.560 \cr
    PU-GCN &0.274&2.974&2.535&0.144&3.713&2.731 \cr
    NePs&0.385 &5.615 &1.642&0.147&8.851&1.925\cr
    PU-SMG &0.296&2.404&2.492&0.155&2.749&2.573 \cr
    {PC}$^2$-PU &0.255&2.504&2.119&0.111&2.806&2.351 \cr
    MAFU &0.280&2.315&1.854&0.156&3.408&1.996 \cr
    
    Grad-PU &0.260&2.462&1.949&0.132&2.421&2.190 \cr
    \midrule
    Ours &\textbf{0.232}&\textbf{1.675}&\textbf{1.338}&\textbf{0.092}&\textbf{1.504}&\textbf{1.544} \cr
    \bottomrule
    \end{tabular}
    
    \caption{Quantitative comparison between our method and  the state-of-the-art methods on the PU-GAN dataset.}
    \label{tab:PUGAN_4X} 
\end{table}

\begin{table}
    \small    
    \centering
    \begin{tabular}{lccc}  
    \toprule 
    \multirow{2}{*}{Methods} &CD$\downarrow$& HD$\downarrow$& P2F$\downarrow$ \cr
    & $10^{-3}$ & $10^{-3}$ & $10^{-3}$\cr
    \midrule
    PU-Net&1.155&15.170&4.834\cr 
    MPU &0.935&13.327&3.511\cr
    PU-GCN &0.585&7.577&2.499\cr
    Dis-PU &0.485&6.145&1.802\cr
    PU-SMG &0.527&4.396&1.647 \cr
    PU-Transformer&0.451&3.843&1.277\cr
    Grad-PU & 0.404&3.732&1.474\cr
    \midrule
    Ours &\textbf{0.371}&\textbf{3.197}&\textbf{1.111}\cr
    \bottomrule
    \end{tabular}
    \caption{Comparison with the state-of-the-art methods on PU1K dataset.}
    \label{tab:PU1K_4X} 
\end{table}

\subsubsection{Evaluation Metrics.}
To evaluate the performance of point cloud upsampling, we use Chamder distance (CD), Hausdorff distance (HD), and point-to-surface distance (P2F) as evaluation metrics following \cite{he2023grad}. 
\subsubsection{Arbitrary-Scale Point Cloud Upsampling.}
We compare our method with previous arbitary-scale upsampling methods including MAFU and Grad-PU on the PU-GAN dataset in Table \ref{tab:PUGAN_arbitrary}. All methods are trained with scale factor $4x$ and are trained for one time. The results show that our method achieves the best performance in various scale factors.
\subsubsection{Baselines.}
We compare the proposed method with several state-of-the-art methods and divide them into fix-scale methods and arbitrary-scale methods according to the scale factor. The fix-scale methods include PU-Net \cite{yu2018pu}, MPU \cite{yifan2019patch}, PU-GAN \cite{li2019pu}, Dis-PU \cite{li2021point}, PU-GCN \cite{qian2021pu}, {PC}$^2$-PU\cite{long2022pc2} and PU-Transformer \cite{qiu2022pu}, and the arbitrary-scale methods include MAFU \cite{qian2021deep}, NePs \cite{feng2022neural}, PU-SMG \cite{dell2022arbitrary}, and Grad-PU \cite{he2023grad}. For a fair comparison, we used the official pre-trained models when available and re-trained the others with the published code with default parameter settings.

\subsection{Evaluation on Synthetic Dataset}
\subsubsection{Results on PU-GAN Dataset.}
As shown in Table \ref{tab:PUGAN_4X}, our method significantly outperforms the baseline methods including PU-Net \cite{yu2018pu}, MPU\cite{yifan2019patch}, PU-GAN \cite{li2019pu}, Dis-PU \cite{li2021point}, PU-GCN \cite{qian2021pu}, NePs \cite{feng2022neural}, PU-SMG \cite{dell2022arbitrary}, PC$^2$-PU \cite{long2022pc2}, MAFU \cite{qian2021deep}, and Grad-PU \cite{he2023grad}, indicating that learning continuous field helps to upsample points accurately on the surface. We also perform a visual comparison in Figure \ref{fig:pugan_compare}, which shows that our method can produce uniform point cloud with fewer holes and outliers. The qualitative and quantitative results for the scale factor $16\times$ show that our method gains larger improvements. The reason behind is that the errors that produced by the local prediction accumulate with the scale factor increasing. Our method will not be affected by these due to the continuous field learning.

\begin{figure}
\centering
\includegraphics[width=1.0\linewidth]{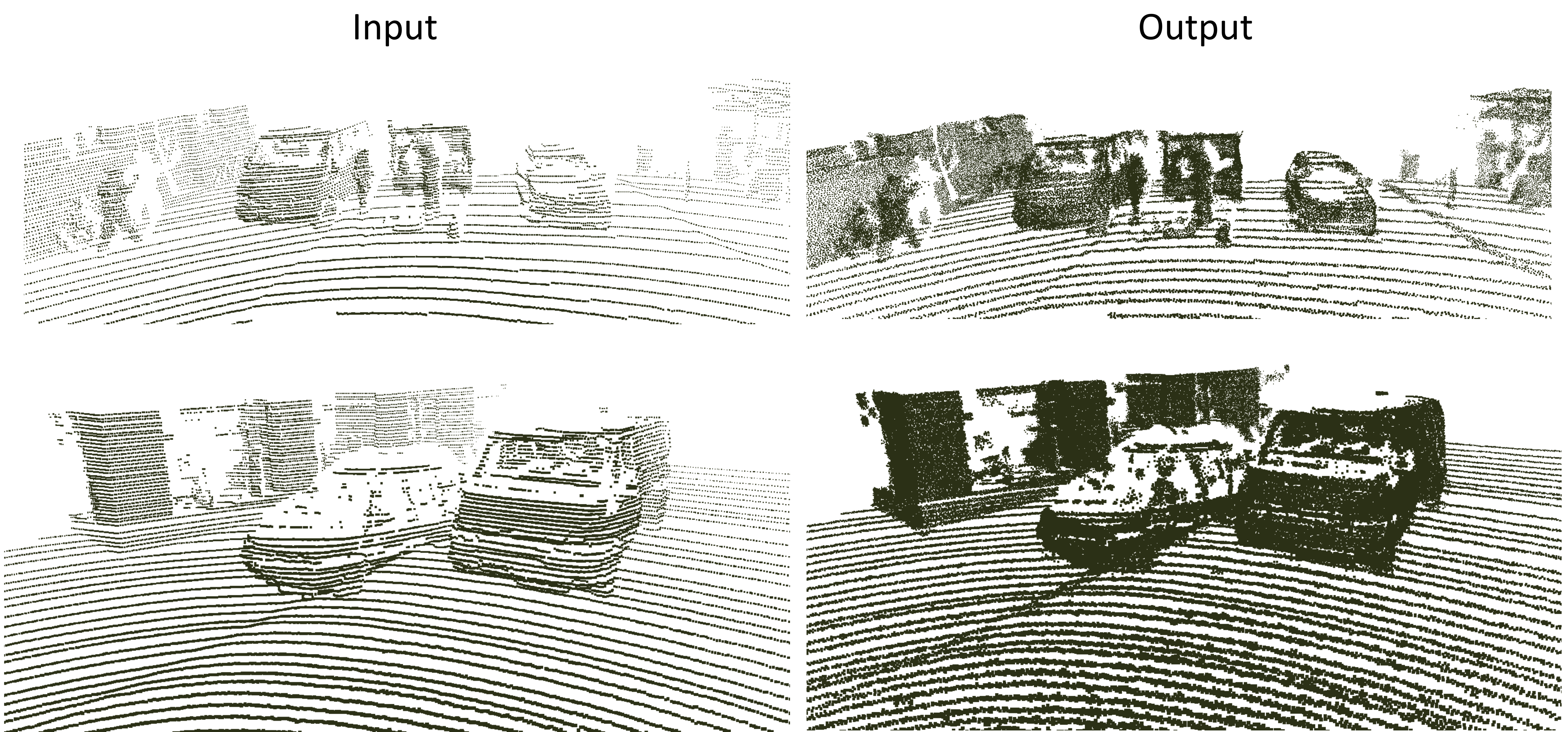}
\caption{The results on KITTI dataset. Our method generates visual-appealing results.}
\label{fig:kitti}
\end{figure}

\begin{table}
    \small    
    \centering
    \begin{tabular}{lccc|ccc}  
    \toprule 
    &\multicolumn{3}{c}{$3\times$ (r=3)}   &\multicolumn{3}{c}{$5\times$ (r=5)}\cr
    \midrule
    \multirow{2}{*}{Methods} &CD$\downarrow$& HD$\downarrow$& P2F$\downarrow$ & CD$\downarrow$& HD$\downarrow$& P2F$\downarrow$ \cr
    & $10^{-3}$ & $10^{-3}$ & $10^{-3}$ & $10^{-3}$ & $10^{-3}$ & $10^{-3}$ \cr
    \midrule
    MAFU&0.555&2.902&1.758&0.252&2.308&1.949\cr
    Grad-PU&0.524&3.050&1.778&0.244&2.447&2.459\cr
    \midrule
    Ours &\textbf{0.448}&\textbf{2.447}&\textbf{1.333}&\textbf{0.193}&\textbf{1.706}&\textbf{1.445} \cr
    \midrule
    &\multicolumn{3}{c}{$7\times$ (r=7)}   &\multicolumn{3}{c}{$13\times$ (r=13)}\cr
    \midrule
    \multirow{2}{*}{Methods} &CD$\downarrow$& HD$\downarrow$& P2F $\downarrow$& CD$\downarrow$& HD$\downarrow$& P2F $\downarrow$\cr
    & $10^{-3}$ & $10^{-3}$ & $10^{-3}$ & $10^{-3}$ & $10^{-3}$ & $10^{-3}$ \cr
    \midrule
    MAFU&0.207&2.593&2.013&0.177&3.212&2.014\cr
    Grad-PU&0.223&2.654&3.325&0.242&4.144&5.703\cr
    \midrule
    Ours &\textbf{0.148}&\textbf{1.534}&\textbf{1.468}&\textbf{0.134}&\textbf{1.878}&\textbf{1.524}\cr
    \bottomrule
    \end{tabular}
    \caption{Quantitative comparison between our method and the state-of-the-art methods with various scale factors.}
    \label{tab:PUGAN_arbitrary} 
\end{table}

\begin{figure}
\centering
\includegraphics[width=1.0\linewidth]{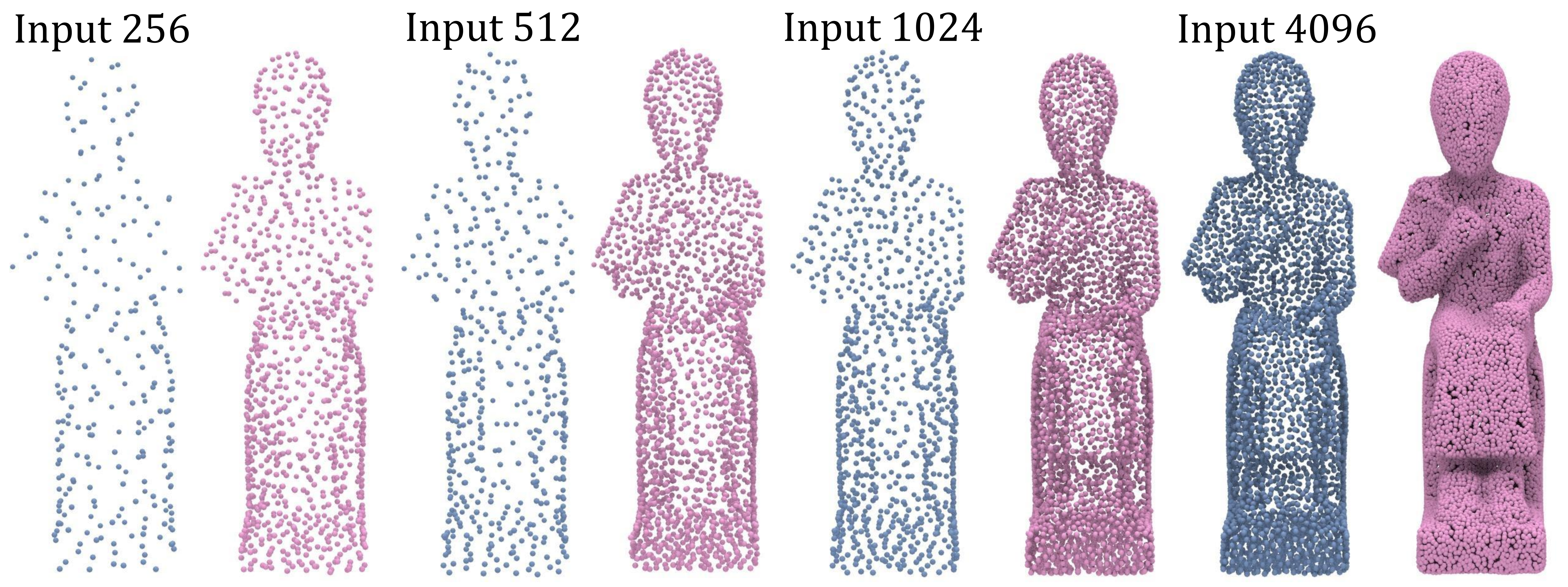}
\caption{$4\times$ results with various input sizes. The blue point clouds are inputs and the red point clouds are corresponding outputs. Our method generates consistent point clouds with various input sizes.}
\label{fig:robust_input_size}
\end{figure}

\begin{figure}
\centering
\includegraphics[width=1.0\linewidth]{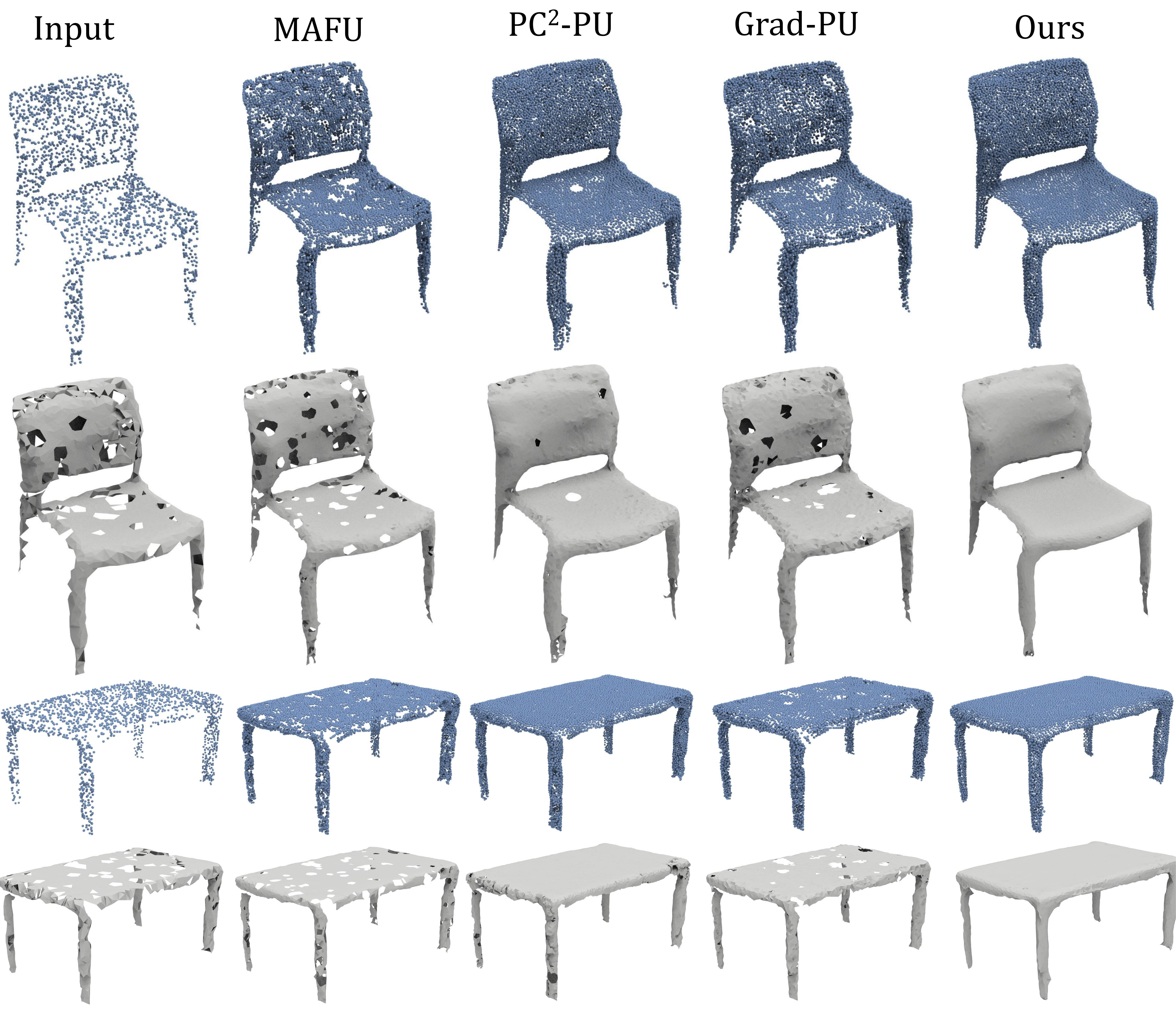}
\caption{The $4\times$ results on ScanObjectNN dataset. The meshes are reconstructed with the BallPivoting algorithm. The results show that our method generates more complete and smooth dense point clouds and meshes.}
\label{fig:scanobjectnn}

\end{figure}

\begin{figure}
\centering
\includegraphics[width=1.0\linewidth]{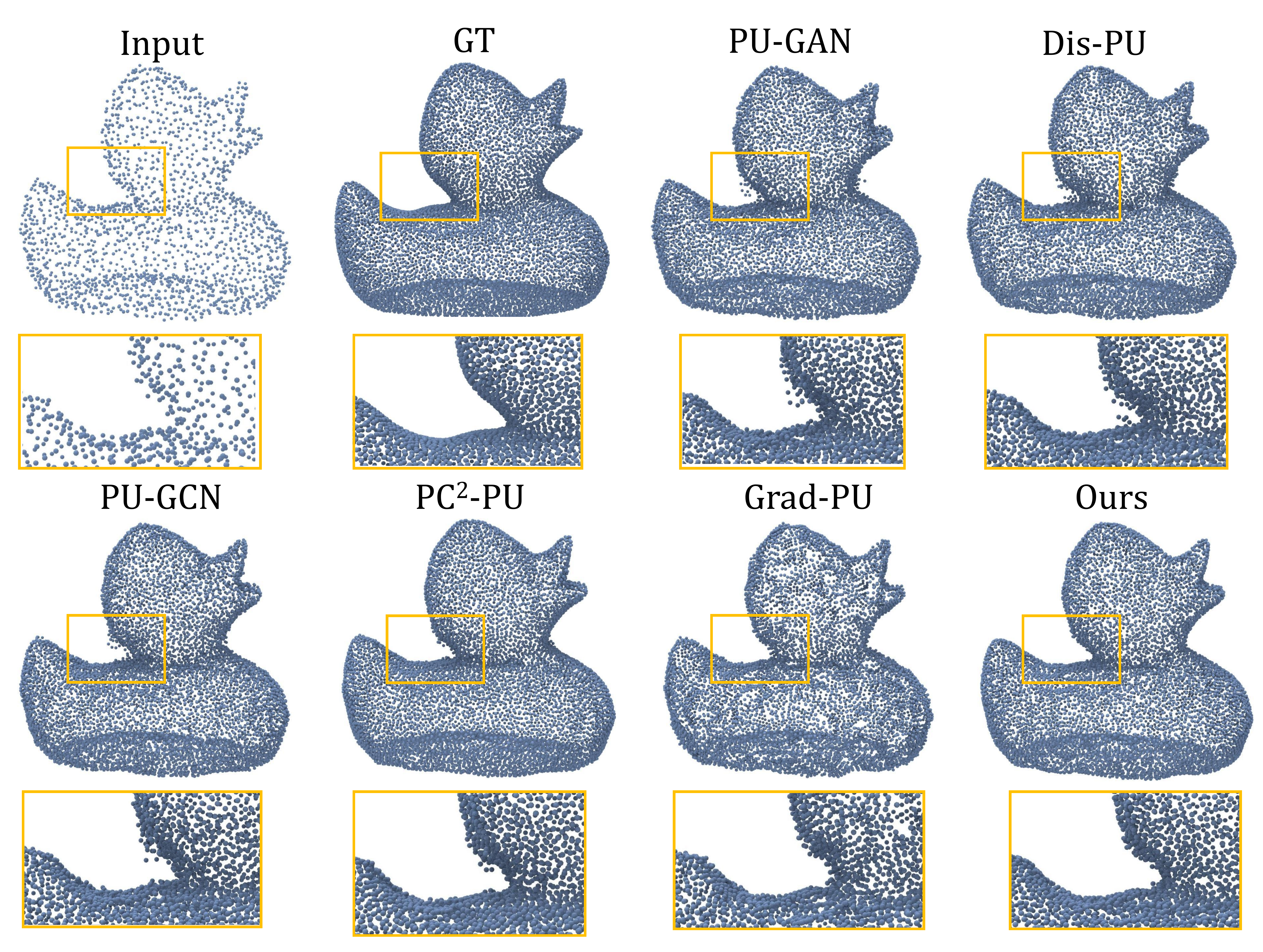}
\caption{$4\times$ results with noise level $\tau=1\%$. The point cloud sampled by our method is cleaner with fewer outliers.}
\label{fig:robust_noise}
\end{figure}

\subsubsection{Results on PU1K Dataset.}
We evaluate our method on the more challenging PU1K dataset. We set the scale factor to $4\times$ and report the comparison results with \cite{yu2018pu, yifan2019patch, qian2021pu, li2021point, qiu2022pu, dell2022arbitrary, he2023grad} in Table \ref{tab:PU1K_4X}, which shows that our method outperforms others on all metrics. 
\begin{table}
    \small    
    \centering
    \resizebox{0.45\textwidth}{!}{
    \begin{tabular}{lcccccc}  
    \toprule 
    Noise Level&\multicolumn{3}{c}{$\tau = 1\%$}   &\multicolumn{3}{c}{$\tau = 2\%$ }\cr
    \cmidrule{2-4} \cmidrule{5-7}
    \multirow{2}{*}{Methods} &CD $\downarrow$ & HD $\downarrow$& P2F  $\downarrow$& CD $\downarrow$& HD$\downarrow$& P2F$\downarrow$ \cr
    & $10^{-3}$ & $10^{-3}$ & $10^{-3}$ & $10^{-3}$ & $10^{-3}$ & $10^{-3}$ \cr
    \midrule
    PU-Net&0.588&6.182&9.842&1.057&9.954&16.282 \cr
    PU-GAN &0.435&7.848&7.300&0.815&9.450&14.246 \cr
    Dis-PU &0.430&6.580&6.954&0.776&8.861&13.934 \cr
    PU-GCN &0.411&5.001&6.963&0.781&8.926&13.730 \cr
    {PC}$^2$-PU &0.369&4.390&5.646&0.733&7.921&12.610 \cr
    Grad-PU &0.423&4.307&6.403&0.730&6.993&11.481 \cr
    \midrule
    Ours &\textbf{0.339}&\textbf{3.089}&\textbf{5.167}&\textbf{0.622}&\textbf{6.485}&\textbf{10.984} \cr
    \bottomrule
    \end{tabular}}
    \caption{Quantitative comparison between our method and the state-of-the-art methods on the PU-GAN dataset with different noise level $\tau$.}
    \label{tab:PUGAN_noise} 
\end{table}

\subsection{Evaluation on Real Scan Dataset}
We validate our method on the real scanned datasets ScanObjectNN \cite{uy2019revisiting} and KITTI \cite{geiger2013vision} with the models trained on the PU-GAN dataset. We only show the qualitative results since the ground truth dense point clouds are unavailable. We utilize BallPivoting \cite{bernardini1999ball} to reconstruct the meshes for the upsampled point clouds on ScanObjectNN and report the results in Figure \ref{fig:scanobjectnn} which show that our method can generate more complete and smooth point clouds and meshes. The testing results on KITTI in Figure \ref{fig:kitti} show that we achieve visual-appealing upsampling results. The comparison with other methods is provided in the supplemental materials.

\subsection{Robustness Test}
\subsubsection{Various Input Size.}
The input sizes of test point cloud in PU-GAN and PU1K are fixed to 2048. To prove that our method is robust to the input size variation, we conduct $4\times$ upsampling experiment on PU-GAN dataset. As shown in Figure \ref{fig:robust_input_size}, our method is not affected by the input sizes and generates consistent results in different point density.

\subsubsection{Additive Noise.} Following previous methods \cite{qian2021pu, long2022pc2, he2023grad}, we conduct $r=4\times$ experiments on PU-GAN dataset to test the robustness of the proposed method. We add Gaussian noise to clean point clouds to generate noisy point clouds at two noise levels ($1\%$ and $2\%$). We compare our method with PU-Net, PU-GAN, Dis-PU, PU-GCN, PC${^2}$-PU and Grad-PU, and all of these methods are trained with the same augmentation strategy of Gaussian noise perturbation. The reported results in Table \ref{tab:PUGAN_noise} show that our method achieves the best performances at both noise levels. Figure \ref{fig:robust_noise} shows that our method can produce point cloud with cleaner surface with fewer outliers, which proves the robustness of our method to noisy input.

\begin{table}
    \small    
    \centering
    \begin{tabular}{l|ccc}  
    \toprule 
    Ablation Settings &CD$\downarrow$& HD $\downarrow$& P2F $\downarrow$ \cr
    \midrule
    Interpolation (GD) & 0.759&5.206&1.753 \cr
    Attention (GD) & 0.678&\underline{4.931}&1.456 \cr
    Interpolation (Global)  &0.714&5.452&1.503 \cr
    OnSurfacePrior&0.719&9.533&2.934\cr
    w/o $L_{\rm{local}}$ &0.874&6.523&1.819\cr
    w/o $L_{\rm{np}}$&\textbf{0.555}&5.720&1.794\cr
    w/o $L_{\rm{sp}}$ &0.746&5.089&\textbf{1.344} \cr
    w/o $L_{\rm{surf}}$ &0.676&5.040&1.541 \cr
    full &\underline{0.637}&\textbf{4.864}& \underline{1.401} \cr
    \bottomrule
    \end{tabular}
    \caption{Results of the ablation study for different experiment settings, with metrics CD($\times 10^{-3}$), HD($\times 10^{-3})$ and P2F($\times 10^{-3}$). Bold and underlined numbers indicate the first and second best performance, respectively.}
    \label{tab:ablation_study}
\end{table}
\subsection{Ablation Studies}
We conduct ablation studies on the subset of the PU1K dataset to show how each module influences the results. We randomly choose about $15\%$ shapes from the testing set as the subset and the results are reported in Table \ref{tab:ablation_study}.

\subsubsection{Module Design.} We first demonstrate the effectiveness of the attention-based module. We adopt the same interpolation operation as \cite{he2023grad} to replace the attention weighting process with the inverse Euclidean distances with neighboring points. At local level, we use the same gradient descent (GD) strategy as Grad-PU. As shown by ``Interpolation (GD)'' and ``Attention (GD)'' in the label, the attention module improves the performance in all metrics.
At global level, we use the interpolation based local distance indicator to guide the learning of continuous field (``Interpolation (Global)''). Comparing with our attention-based indicator (``full''), we find a drop in performance, which shows that the attention-based module plays a positive role in the learning of the global implicit field.

As shown by ``Interpolation (GD)'' vs ``Interpolation (Global)'' and ``Attention (GD)'' vs ``full'', the global implicit field strategy performances better than the gradient descent strategy in almost all metrics, which shows the global implicit field helps to generate point clouds with higher quality.

We conduct the experiment to compare our method with OnSurfacePrior \cite{ma2022reconstructing} that also has the combination of local indicator and global learning as shown by ``OnSurfacePrior'' in Table \ref{tab:ablation_study}. The results demonstrates that our designs are more adaptive to the point cloud upsampling. 

\subsubsection{Loss Function.}  We remove each loss term from the full setting to show how it influences the final result. As shown by ``w/o $L_{\rm{local}}$'', the performance degenerates dramatically without the local distance indicator. As shown by ``w/o $L_{\rm{np}}$'', the nearest point loss increases the CD, but it improves the other metrics significantly. The reason for the metric increase is that the nearest point on the sparse point is not exact the nearest point on the underlying surface, which leads to approximation errors. The third row ``w/o $L_{\rm{sp}}$'' shows that the shortest path loss improves the CD and HD, but leads to an increase on P2F for the introduced noises. The surface loss can improve the performance on all metrics.

\section{Conclusion}
In this paper, we present a novel method to learn an unsigned distance field with a local distance indicator for arbitrary-scale point cloud sampling. Instead of upsampling the point sets at the local level as existing methods, we propose to represent the entire point cloud in the global implicit field to keep the consistency among local patches. We first design an attention-based local distance indicator that predicts the distance from a query location to the surface of the local patch, and then utilize it to guide the learning of implicit field for the point cloud.  Extensive quantitative and qualitative comparisons on synthetic data and real scans demonstrate that our method outperforms prior state-of-the-art methods.

\section{Acknowledgments}
This work was supported by National Key R\&D Program of China (2022YFC3800600), the National Natural Science Foundation of China (62272263, 62072268), and in part by Tsinghua-Kuaishou Institute of Future Media Data.

\bibliography{aaai24}

\end{document}